\documentclass[sigconf]{acmart}

\usepackage[T1]{fontenc}
\usepackage{graphicx}
\usepackage{booktabs}
\usepackage{mwe}
\usepackage{hyperref}
\usepackage{url}
\usepackage{algorithmic}
\usepackage{algorithm}
\usepackage{graphicx}
\usepackage{booktabs}
\usepackage{multirow}
\usepackage{wrapfig}
\usepackage{enumitem}
\usepackage{amsfonts}
\usepackage{amsmath}
\usepackage{xcolor}
\usepackage[normalem]{ulem}

\usepackage{amsmath,amsfonts,bm}

\def\eqref#1{equation~\ref{#1}}

\def\1{\bm{1}}

\DeclareMathAlphabet{\mathsfit}{\encodingdefault}{\sfdefault}{m}{sl}
\SetMathAlphabet{\mathsfit}{bold}{\encodingdefault}{\sfdefault}{bx}{n}

\def\sA{{\mathbb{A}}}

\newcommand{\E}{\mathbb{E}}

\DeclareMathOperator*{\argmax}{arg\,max}

\DeclareMathOperator*{\minimize}{minimize}

\newenvironment{ttable}[2][c]
 {%
  \begin{tabular}[#1]{|@{}c@{}|}
  \begin{tabular}[#1]{#2}
 }
 {%
  \end{tabular}
  \end{tabular}
 }

\AtBeginDocument{%
  \providecommand\BibTeX{{%
    \normalfont B\kern-0.5em{\scshape i\kern-0.25em b}\kern-0.8em\TeX}}}

\setcopyright{acmcopyright}
\copyrightyear{2024}
\acmYear{2024}
\acmDOI{10.1145/nnnnnnn.nnnnnnn}

\acmConference[AIML Systems 2024]{The Fourth International Conference on Artificial Intelligence and Machine Learning Systems (AIMLSystems 2024)}{October 08--11, 2024}{Baton Rouge}
\acmBooktitle{The Fourth International Conference on Artificial Intelligence and Machine Learning Systems (AIMLSystems 2024), October 08--11, 2024, Lousiana State University, Baton Rouge, USA} 
\acmISBN{979-8-4007-1161-9/24/10}

\begin{document}

\title{ILAEDA: An Imitation Learning Based Approach for Automatic Exploratory Data Analysis}

\author{Abhijit Manatkar}
\email{abhijit.manatkar@students.iiit.ac.in}
\affiliation{%
  \institution{Machine Learning Lab, KCIS, IIIT Hyderabad}
  \city{Hyderabad}
  \state{Telangana}
  \country{India}
  \postcode{500032}
}

\author{Devarsh Patel}
\email{devarsh.patel@students.iiserpune.ac.in}
\affiliation{%
  \institution{IISER, Pune}
  \city{Pune}
  \state{Maharashtra}
  \country{India}
  \postcode{411008}
  }

\author{Hima Patel}
\email{himapatel@in.ibm.com}
\affiliation{%
  \institution{IBM Research, India}
  \city{Bangalore}
  \state{Karnataka}
  \country{India}
  \postcode{560045}
}

\author{Naresh Manwani}
\email{naresh.manwani@iiit.ac.in}
\affiliation{%
  \institution{Machine Learning Lab, KCIS, IIIT Hyderabad}
  \city{Hyderabad}
  \state{Telangana}
  \country{India}
  \postcode{500032}
}



\begin{abstract}
  Automating end-to-end Exploratory Data Analysis (AutoEDA) is a challenging open problem, often tackled through Reinforcement Learning (RL) by learning to predict a sequence of analysis operations (FILTER, GROUP, etc). Defining rewards for each operation is a challenging task and existing methods rely on various \emph{interestingness measures} to craft reward functions to capture the importance of each operation. In this work, we argue that not all of the essential features of what makes an operation important can be accurately captured mathematically using rewards. We propose an AutoEDA model trained through imitation learning from expert EDA sessions, bypassing the need for manually defined interestingness measures. Our method, based on generative adversarial imitation learning (GAIL), generalizes well across datasets, even with limited expert data. We also introduce a novel approach for generating synthetic EDA demonstrations for training. Our method outperforms the existing state-of-the-art end-to-end EDA approach on benchmarks by upto 3x, showing strong performance and generalization, while naturally capturing diverse interestingness measures in generated EDA sessions.
\end{abstract}




\maketitle

\section{Introduction}
Exploratory Data Analysis (EDA) stands as a pivotal step within any data science pipeline, offering an effective means to comprehend datasets by uncovering patterns, anomalies, and critical insights. Traditionally, this process has heavily relied on human intuition, leading to a trial-and-error exploration guided by the analyst's domain expertise. However, recent advancements have sought to automate this process through AutoEDA methods \cite{sarawagi1998discovery, Razmadze2022SelectingSF, somech2019predicting, bar2020automatically, milo2016react}, aiming to provide analysts with initial insights into datasets automatically, which serve as a springboard for further exploration.

Some of these efforts have sought to automate the entire EDA process using deep reinforcement learning (RL)-based AutoEDA systems \cite{bar2020automatically}  \cite{personnaz2021balancing}. These systems treat AutoEDA as a sequential decision-making problem, with RL algorithms trained to predict a sequence of EDA operations. The rewards for such RL systems are heuristics/rules. However, such approaches face significant challenges:
\begin{enumerate}
    \item Defining appropriate reward functions that encapsulate all the relevant aspects of a dataset is inherently complex. Often, the reward function used is insufficient to capture all the interesting dimensions of the data.
    \item For a given dataset, previous expert data analysis sessions provide major hints about the usefulness of certain reward functions. One has to analyze these expert sessions to get those hints. 
    \item Precise dataset-specific reward function definitions are required to learn meaningful policies in RL-based systems like the one presented by \cite{bar2020automatically}. Defining detailed dataset-specific reward functions requires a good deal of data analysis to already have been done, a requirement which contradicts the motivations behind building an AutoEDA system.
\end{enumerate}

To address these challenges, we propose \textbf{ILAEDA}, an imitation learning-based framework for AutoEDA. Unlike reward-based methods, ILAEDA learns directly from expert EDA sessions, eliminating the need for handcrafted reward functions and dataset-specific analyses. Our approach learns to imitate a human when shown a few expert sessions. To make sure that we don't overfit the model to the available expert sessions, we introduce a novel approach to automatically generate synthetic EDA demonstrations that are used for training. This helps our system to offer better generalizability across datasets along with the ability to capture diverse aspects of effective EDA sessions, such as diversity and coherence. {\bf Key contributions} of our work include:
\begin{enumerate}
\item Introducing ILAEDA, an imitation learning framework for AutoEDA, which removes the reliance on handcrafted rewards and dataset-specific analyses.
\item Showcasing ILAEDA's enhanced generalizability across datasets and its capacity to capture various facets of effective EDA sessions. As our approach trains a model from expert demonstrations, a diverse set of demonstrations allows for generalization across different datasets. 
\item Providing a comprehensive analysis of our results, demonstrating ILAEDA's effectiveness even with minimal expert demonstrations. Experimental results show that the sessions generated by our model automatically capture multiple different notions of what makes for a good EDA session (diversity, coherence, readability, etc.) without using reward scores.
\end{enumerate}

\section{Related Work}
\label{sec:related-work}

\subsection{AutoEDA Methods}
In recent years, several methods have attempted to automate different aspects of the EDA process. A set of next-step recommendation systems make use of pre-defined notions of interestingness to quantify the output of a specific EDA operation, e.g. for data-visualization \cite{vartak2015seedb}, for finding interesting data tuple-subsets or data cube subsets \cite{sarawagi1998discovery} \cite{sarawagi2000user} \cite{drosou2013ymaldb} \cite{Razmadze2022SelectingSF},  OLAP drill-down \cite{joglekar2017interactive}, and data summaries \cite{singh2016dbexplorer}. Log-based EDA recommender systems, such as \cite{aligon2015collaborative} \cite{eirinaki2013querie} \cite{yang2009recommending}, use a collection of the previous exploratory sessions of the same or different users to generate recommendations for the next exploratory operation. Systems in \cite{milo2016react} \cite{milo2018next} utilize session logs to retrieve similar prefixes, generate candidate \emph{next-actions}, and convert them into concrete recommendations of EDA operations based on interestingness measures. The approach in \cite{somech2019predicting} recommends next action using a kNN-based classifier on user exploration logs. This classifier predicts the best interestingness measure that could capture user's interest at every step of the EDA session. EDA can also be modeled as a meta-learning problem \cite{Cao2022LearnTE}.

The approach in \cite{Chanson2022AutomaticGO} generates comparison queries to highlight insights in a dataset. In \cite{Ma2021MetaInsightAD}, authors propose a novel formulation of basic data patterns to facilitate extraction of insights from a dataset. \cite{Ma2023InsightPilotAL} present a Large Language Model 
\cite{touvron2023llama} based system for extracting insights from a dataset and presenting them to the user through a natural language interface.

RL based AutoEDA systems are described in \cite{bar2020automatically} \cite{Personnaz2021DORATE} \cite{Lipman2023ATENAPROGP}. 
ATENA \cite{bar2020automatically} automatically generates EDA sessions for a given tabular dataset. 
In this approach, EDA is formulated as a sequential decision-making problem. 
A deep reinforcement learning algorithm is used to learn a policy to generate a sequence of EDA operations which maximize the defined reward function. Authors of \cite{Lipman2023ATENAPROGP} describe a system built with ATENA as a base, augmenting it with a novel action space formulation and user-specified constraint compliance rewards.

\subsection{Imitation Learning}

In this setup, we do not have access to the reward structure. 
We want to learn the optimal policy without knowing the rewards. Behavioral cloning \cite{10.5555/3304652.3304697} \cite{Ross2010EfficientRF}, Imitation learning \cite{NEURIPS2019_c8d3a760}\cite{ho2016generative} \cite{Reddy2019SQILIL} \cite{Garg2021IQLearnIS}, Inverse RL 
\cite{10.5555/1620270.1620297} are some of the techniques which are used to learn an optimal policy when we do not know the reward structure. All these approaches require expert trajectories to learn an optimal policy.

Generative Adversarial Imitation Learning (GAIL) \cite{ho2016generative} treats imitation learning as a min-max adversarial problem. The method describes two modules - a discriminator $D$ and a policy $\pi$ with opposing objectives:
\begin{enumerate}
    \item Given any state $s$, the objective of the policy is model the distribution $\pi(a | s)$ to mimic the expert demonstrations as closely as possible.
    \item Given a pair of state and action taken in the state $(s,a)$, the objective of the discriminator is to differentiate between $(s,a)$ pairs coming from an expert session and the policy. For any $(s,a)$, $D(s,a) \in [0, 1]$,  where $D(s,a) = 1$ indicates that $(s,a)$ comes from expert demonstrations and $D(s,a) = 0$ indicates that it is a generated pair.
\end{enumerate}
The two modules are optimized in a procedure that alternates between optimizing $D$ and optimizing $\pi$ until $D$ is no longer able to distinguish between state-action pairs coming from expert demonstrations and state-action pairs coming from $\pi$. 
GAIL has been shown to effectively imitate expert demonstrations even when the number of demonstrations is very low, which makes it an attractive candidate for the end-to-end AutoEDA problem as the availability of expert demonstrations is generally quite low.

\begin{table}[h]
    \centering
    \footnotesize
    \resizebox{\linewidth}{!}{
    \begin{ttable}{l|r}
    \toprule
    \textbf{Action} & \textbf{Measure} \\
    \hline \\
    GROUP highest\_layer AGGREGATE COUNT \\ packet\_number & A-INT, Diversity\\
    \hline \\
    BACK & - \\
    \hline \\
    GROUP eth\_src AGGREGATE COUNT \\ 
    packet\_number &  A-INT, Readability\\
    \hline \\
    GROUP ip\_src AGGREGATE COUNT \\ 
    packet\_number &  Peculiarity\\
    \hline \\
    FILTER info\_line CONTAINS \\
    Echo (ping) reply &  Diversity, Coherence\\
    \hline \\
    BACK & - \\
    \hline \\
    BACK & - \\
    \hline \\
    BACK & - \\
    \hline \\
    FILTER highest\_layer NEQ  ICMP & Coherence\\
    \hline \\
    GROUP tcp\_srcport AGGREGATE COUNT packet\_number & Diversity\\
    \hline \\
    GROUP ip\_src AGGREGATE COUNT \\ 
    packet\_number & Coherence, Peculiarity\\
    \hline \\
    FILTER ip\_src NEQ  192.168.1.122 &  Coherence \\ 
    \bottomrule
    \end{ttable}
    }
    \caption{\label{tab:interesting1}Expert EDA session and the interestingness measure maximized by each operation in the analysis sequence. Measures with relative score greater than 0.8 are mentioned}
\end{table}

\subsection{Interestingness Measures} 
Various interestingness measures have been studied in literature. Some of them have been used as rewards in RL based systems and they are also useful for analyzing EDA operations. We present a brief summary of some of the interestingness measures studied in literature, which we use in later sections.\footnote{For a detailed description, refer to Appendix \ref{appendix:interestingness-measures}} (a) \textbf{A-INT.} This interestingness score \cite{bar2020automatically} is used as a reward function that rewards display which are compact in case of GROUP operations, and displays with high deviation from the previous display in case of FILTER operations. (b) \textbf{Diversity score.} This metric favors a display that highlights parts of the dataset that are different from those seen in any of the previous displays in the session so far \cite{bar2020automatically}. (c) \textbf{Readability score.} This metric rewards highly compact displays \cite{chandola2007summarization} with few rows as they are considered more readable and are given a higher reward. (d) \textbf{Peculiarity score.} This score helps quantify anomalous patterns by favoring display with a high difference from the initial display at the start of the analysis \cite{vartak2015seedb}. (e) \textbf{Coherence score.} This score is a highly detailed metric determined by a set of handcrafted rules which assign each display a penalty or reward based on whether the operation performed at the current step is coherent with previous operations \cite{bar2020automatically}.


\section{Problem Setup}


\label{sec:analysis-intro}
Our proposed approach attempts to tackle the end-to-end AutoEDA problem using imitation learning. In this section, we describe the problem setup and illustrate some of the challenges with reward-based approaches for this problem. 

The input is a tabular dataset $\mathcal{D}$ with a set of attributes (columns) $\sA$ and the desired output is a sequence of analysis operations which will form an \emph{EDA session}. This problem can be modeled as a Markov Decision Process $\langle \mathcal{S}, \mathcal{A}, R, \mathcal{P} \rangle$ where $\mathcal{S}, \mathcal{A}$ are the state and action space respectively, $R: \mathcal{S} \times \mathcal{A} \rightarrow \mathbb{R}$ is the reward function, and $\mathcal{P}: \mathcal{S} \times \mathcal{A} \rightarrow \mathcal{S}$ is a deterministic transition function. Let $\Pi$ be the space of all stationary stochastic policies which return a distribution over actions in $\mathcal{A}$ given a state in $\mathcal{S}$. The goal is to learn an optimal policy $\pi^\ast \in \Pi$ which maximizes a discounted cumulative sum of rewards for a fixed horizon $T$, i.e.
$\pi^\ast = \argmax_{\pi}{\E_{\pi}[\sum_{t = 0}^{T} \gamma^t R(s_t, a_t)]}$,
where $s_t \in \mathcal{S}$ is the state at time-step $t$ given by $s_t = \mathcal{P}(s_{t-1}, a_{t-1})$, $a_t \sim \pi(\cdot | s_t)$ and $\gamma \in [0, 1]$ is the discounting factor. 

\subsection{State Space}
State space \cite{bar2020automatically} is described as follows. An EDA session begins with an initial display $d_0$ of the dataset $\mathcal{D}$. At each subsequent timestep $t$, an action $a_t \in \mathcal{A}$, which is an EDA operation, is applied to display $d_t$ to obtain display $d_{t+1}$. For each display $d_t$, corresponding state $s_t \in \mathcal{S}$ is represented as $s_t = \text{Encode}(d_{t-2}) \oplus \text{Encode}(d_{t-1}) \oplus \text{Encode}(d_{t})$,
where $\oplus$ is the vector concatenation operator and Encode($d_t$) encodes a display into a vector by extracting the following features. (a) For each attribute $A \in \mathbf{A}$ in the display, we include its entropy, number of distinct values, and the number of null values. (b) For each attribute $A \in \mathbf{A}$, we include a feature to determine whether that column is grouped, aggregated, or neither. (c) Three global features: number of groups, the mean size of groups, and variance of size of groups.
If less than three displays have been seen, Encode($d_{t-2}$) and Encode($d_{t-1}$) are replaced with 0 vectors of appropriate sizes as required.

\begin{figure}
  \centering
  \includegraphics[width=.45\textwidth]{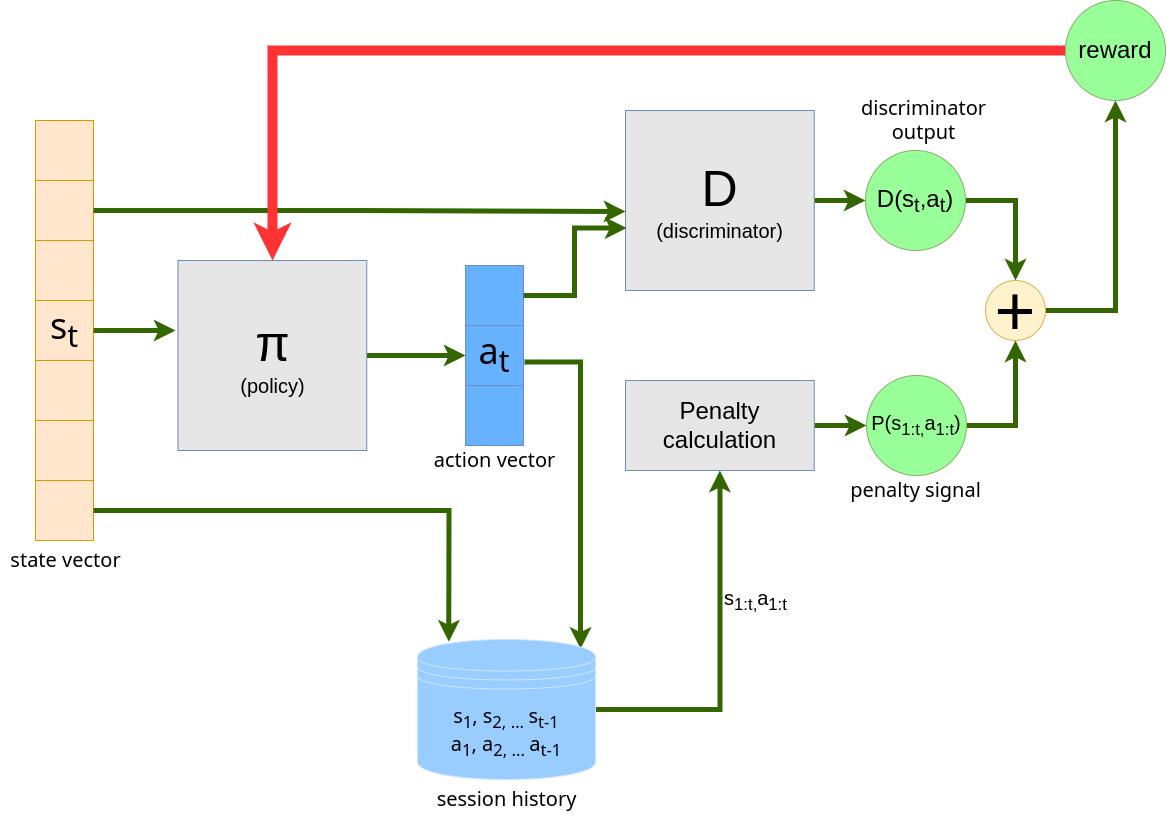}
  \caption{ILAEDA System Architecture}
  \label{system-flow-diagram}
\end{figure}

\subsection{Action Space}

The action space $\mathcal{A}$, in addition to basic BACK and STOP actions for directing the flow of the analysis, consists of parameterized FILTER and GROUP operations with possible parameters drawn from the columns of the dataset $\mathbf{A}$ and their distributions as described below:
\begin{enumerate}
    \item \texttt{GROUP(grp\_col, agg\_col, agg\_func)}: This action groups on the \texttt{grp\_col} $\in \mathbf{A}$ and aggregates the \texttt{agg\_col} $\in \mathbf{A}$ using the \texttt{agg\_func}, which can be one of \{ \texttt{SUM, COUNT, MEAN, MIN, MAX} \}.
    \item \texttt{FILTER(filter\_col, filter\_func, filter\_term)}: This action filters the dataset by using the comparison operator $\texttt{filter\_func} \in \{ =, \neq, \texttt{CONTAINS}, \texttt{STARTS\_WITH}, \texttt{ENDS\_WITH} \}$ and \texttt{filter\_term}, which (numerical or textual) appears from within the $\texttt{filter\_col} \in \mathbf{A}$.
    \item \texttt{BACK()}: Allows to take a step backward and return to the previous display in the ongoing analysis session.
    \item \texttt{STOP()}: Allows the signaling of the end of the current analysis session.
\end{enumerate}

\subsection{Issues With Reward-Based Modeling}

ATENA \cite{bar2020automatically} uses hand-crafted rewards based on interestingness measures. The reward function used is given by:
\begin{equation}
    \label{eq:atena-reward}
    R(s_t, a_t) = r_{int}(s_t, a_t) + \lambda_1 r_{div}(s_t, a_t) + \lambda_2 r_{coh}(s_t, a_t) 
\end{equation}
Here, $r_{int}, \;r_{div}$ are the A-INT and Diversity score respectively (refer Section \ref{sec:related-work}). $r_{coh}$ is the output of a binary classifier trained using rules based on the Coherence score. $\lambda_1, \lambda_2$ are constants that are tuned to calibrate the reward. 
As described before, some of the rules defined for the Coherence reward include:
\begin{enumerate}
    \item Filtering/Grouping on a specific set of filterable or groupable columns is rewarded, and specific non-filterable or non-groupable columns are penalized.
    \item Using a specific set of operators to filter the \texttt{info\_line} column is rewarded.
    \item Grouping on column \texttt{highest\_layer} on a filtered display is penalized.
\end{enumerate}
Evidently, these rules are highly dataset-specific and require a great deal of familiarity with the dataset. This is one major drawback of reward based model.

To understand futher shortcomings of reward-based modeling, we analyze an expert EDA session using the interestingness measures described in Section \ref{sec:related-work}. Table \ref{tab:interesting1} shows sequence of operations from an EDA session performed by an expert on the dataset mentioned above. At each step of the session, we calculate the scores of interestingness measures and examine which of them are optimized at each step. The scores are normalized \cite{somech2019predicting} in the range of $[0,1]$ for comparison. At every step, measures with a normalized score exceeding 0.8 are shown. We observe that:
\begin{enumerate}
    \item Within the same expert session, every step optimizes for different combinations of interestingness measures.
    \item  Some actions optimize the Peculiarity and Readability scores, which are not part of the reward definition in \ref{eq:atena-reward}.
\end{enumerate}

Consistent observations across multiple expert sessions reveal the notable limitation that handcrafted rewards often fail to capture the full spectrum of intriguing data aspects. Moreover, our analysis suggests that this limitation becomes more pronounced when considering a broader array of measures. These identified drawbacks serve as a compelling motivation for us to propose an alternative approach to learning an optimal policy for ILAEDA, which aims to mitigate these issues.

\section{ILAEDA: Proposed Methodology}
\label{sec:Method}

The overall architecture of the proposed approach ILAEDA is provided in Figure~\ref{system-flow-diagram}. Our main goal is to learn a policy directly from the expert EDA sessions (from here on, also referred to as \textit{expert trajectories}) using Imitation Learning, allowing the system to learn to imitate a human expert.
We have a set of expert trajectories $\mathcal{E} = \{\tau_1, \tau_2, \dots \tau_N\}$ where each $\tau_i$ is a sequence of state-action pairs $\{(s_1, a_1), (s_2, a_2) \dots (s_T, a_T)\}$ where each $s_i \in \mathcal{S}$ and $a_i \in \mathcal{A}$. We assume that expert actions come from an underlying expert policy $\pi_E$, i.e.
$a_t \sim \pi_E(\cdot | s_t),\; \forall \, (s_t, a_t) \in \tau,\; \forall \tau \in \mathcal{E}$. ILAEDA uses three neural networks: Policy $\pi_\theta$, Value $V_\phi$ and Discriminator $D_w$ parameterized by $\theta, \phi, w$ respectively. 
The policy and discriminator networks undergo training at alternate steps. The discriminator is trained to minimize the loss: 
\begin{equation}
    L_D = \E_{\pi_\theta} [ \log( D_w(s, a) )] + \E_{\pi_E} [ \log(1 - D_w(s, a) )]
\end{equation}

In the first term, the expectation is over a mini-batch of $(s,a)$ pairs collected by environment interaction using the current policy $\pi_\theta$. 
A set of trajectories $\mathcal{T} = \{ \tau^\theta_1 \dots \tau^\theta_N \}$ is obtained where each $\tau^\theta_i = \{(s_1, a_1) \dots (s_T, a_T)\}$ consists of $(s,a)$ pairs such that action $a \sim \pi_\theta(\cdot | s)$. In the second term, the expectation is over an equal sized mini-batch of $(s,a)$ pairs randomly picked from $\tau \in \mathcal{E}$. The policy and value networks are trained using Proximal Policy Optimization (PPO) \cite{schulman2017proximal} on a mini-batch of 4-tuples $(r_t, s_t, a_t, s_{t+1})$ where $r_t = - \log (1 - D_w(s_t, a_t)) + P(s_{1:t}, a_{1:t})$. The first term is the imitation reward, based on the discriminator output as in GAIL \cite{ho2016generative}. The second term is our penalty term that we describe below.

Half of the mini-batch is composed of expert demonstrations $(s_t, a_t, s_{t+1})$ obtained from $\tau \in \mathcal{E}$, each of which is appended with a reward term $r_t$ derived from the output of the discriminator as described above. The other half is composed of $(s_t, a_t, s_{t+1})$ obtained from $\tau^\theta \in \mathcal{T}$ with $r_t$ appended as above. The supplementary material provides a detailed algorithmic description of the overall method. Three key aspects of training of ILAEDA are as follows.

\subsection{Penalties for Incoherent Action Sequences}
An important feature of ILAEDA is that it augments the reward term 
with a penalty term for incoherent sequences of actions. Some such incoherent sequence of actions are as follows.
(a) Model takes a BACK action at the beginning of the analysis, (b) model consecutively repeats the same FILTER/GROUP action, (c) model alternates between a FILTER/GROUP action and BACK action.

A sequence of operations is typically helpful when it consists of at least 2-3 continuous FILTER/GROUP operations before the steps are retraced using BACK. Taking a single FILTER/GROUP action and immediately retracing with a BACK action, and continuing to do so repeatedly signals a lack of confidence in the chosen path, and hence an uncertainty in the actions taken. To discourage the learning of these behaviours, we use the following reward function,
\begin{equation}
    \label{eq:gail-penalty-reward}
    r_t = - \log (1 - D_w(s_t, a_t)) + P(s_{1:t}, a_{1:t})
\end{equation}
where $P(s_{1:t}, a_{1:t})$ is an incoherence penalty defined as follows:
\begin{equation}
    P(s_{1:t}, a_{1:t})=
        \begin{cases}
            -1.0 & \text{if } a_t = \texttt{BACK} \text{ and } t = 1 \\
            -1.0 & \text{if } a_t \neq \texttt{BACK} \text{ and } a_t = a_{t-1} \\
            -1.0 \times l & \parbox[t]{4.5cm}{\text{if } $a_t = a_{t-2} \dots = a_{t-2l} = \texttt{BACK} \neq a_{t-2(l+1)}$ \text{ and } $a_{t-1}, a_{t-3} \dots ,a_{t-2l-1} \in \{\texttt{FILTER}, \texttt{GROUP}\}$ \text{ and } $l > 1$} \\
            0.0 & \text{otherwise}
        \end{cases}
\end{equation}

\begin{table*}[htb]
\centering
\small
\begin{tabular}{lcccccc}
\toprule
\textbf{Dataset} & \textbf{Method} & \textbf{Precision} & \textbf{TBLEU-1} & \textbf{TBLEU-2} & \textbf{TBLEU-3} & \textbf{EDA-Sim} \\
\midrule
\multirow{3}{*}{CS 1} & BC & 0.2117 & 0.1421 & 0.0826 & 0.0519 & 0.2496 \\
                      & ATENA & 0.1855 & 0.1855 & 0.1377 & 0.0625 & 0.2704 \\
                      & ILAEDA & \textbf{0.3750} & \textbf{0.3333} & \textbf{0.2041} & \textbf{0.0841} & \textbf{0.2950} \\
\midrule
\multirow{3}{*}{CS 2} & BC & 0.3287 & 0.1497 & 0.0726 & 0.0497 & 0.3491 \\
                      & ATENA & 0.2340 & 0.2325 & 0.1873 & \textbf{0.1182} & 0.2682 \\
                      & ILAEDA & \textbf{0.4000} & \textbf{0.2857} & \textbf{0.2182} & 0.1060 & \textbf{0.3900} \\
\midrule
\multirow{3}{*}{CS 3} & BC & \textbf{0.1614} & 0.1235 & 0.0456 & \textbf{0.0393} & 0.2728 \\
                      & ATENA & 0.1153 & 0.1122 & \textbf{0.0550} & 0.0320 & 0.2462 \\
                      & ILAEDA & 0.1429 & \textbf{0.1314} & 0.0449 & 0.0359 & \textbf{0.3497} \\
\midrule
\multirow{3}{*}{CS 4} & BC & 0.1950 & 0.1474 & 0.0771 & 0.0472 & \textbf{0.3378} \\
                      & ATENA & 0.1929 & 0.1929 & 0.1451 & 0.0708 & 0.3017 \\
                      & ILAEDA & \textbf{0.7500} & \textbf{0.3333} & \textbf{0.2041} & \textbf{0.0841} & 0.2051 \\
\midrule
\multirow{3}{*}{SD 6} & BC & \textbf{0.9167} & 0.1428 & 0.1092 & 0.0612 & 0.2929 \\
                      & ATENA & 0.1111 & 0.0515 & 0.0173 & 0.0132 & 0.1539 \\
                      & ILAEDA & 0.4286 & \textbf{0.4286} & \textbf{0.1816} & \textbf{0.0669} & \textbf{0.5647} \\
\midrule
\multirow{3}{*}{SD 7} & BC & 0.7071 & 0.2190 & 0.1520 & 0.0932 & 0.3505 \\
                      & ATENA & 0.1111 & 0.0429 & 0.0143 & 0.0114 & 0.1265 \\
                      & ILAEDA & \textbf{0.8333} & \textbf{0.5333} & \textbf{0.4781} & \textbf{0.3852} & \textbf{0.5536} \\
\bottomrule
\end{tabular}
\caption{\label{tab:benchmarkscores}Comparison of ILAEDA with baselines on Cyber Security (CS 1-4) and Synthetic Datasets (SD 6-7)}
\end{table*}

\subsection{Initialization of Policy with Behavioral Cloning}
We initialize the policy network using a behavioral cloning objective. We observe that such initialization results in stronger final performance of the learnt model. Our reasoning for this finding is that behavioural cloning provides a \emph{warm start} to the policy network compared to random initialization of parameters. As very few expert demonstrations are available for imitation, pre-training with behavioural cloning is valuable in initializing the policy at a point in parameter space which is likely to converge to the optimum faster. The pre-training stage seeks to minimize the following objective:
\begin{equation}
    \minimize_\theta \E_{(s,a) \in \mathcal{E}} \left[ -\log \pi_\theta(a | s) + || \theta ||_2 \right]
\end{equation}
where $(s,a)$ pairs come from trajectories $\tau_E \in \mathcal{E}$. We conduct an ablation study to examine the effects of this initialization choice.


\subsection{Proximal Policy Optimization Updates}
In ILAEDA, we train the policy network using Proximal Policy Optimization (PPO) \cite{schulman2017proximal}. 
Specifically, we use the PPO-clip update given as:
\begin{equation}
    \theta_{t+1} = \argmax_{\theta} \E_{T \sim \mathcal{B}} \left[ \min \left( \frac{\pi_{\theta}(a | s)}{\pi_{\theta_t}(a | s)}A^t(T), g \left( \epsilon, A^t(T) \right) \right) \right]
\end{equation}
Here $\pi_{\theta_{t}}$ is the policy at the $t$ th time step. We wish to update parameters $\theta_{t}$ to $\theta_{t+1}$ such that expected reward is maximized while making sure that $\pi_{\theta_{t+1}}$ is not too far from $\pi_{\theta_{t}}$. $\mathcal{B}$ is a set of transitions $T = (s, a, r, s')$ sampled from environment interaction with policy $\pi_{\theta_{t}}$. $A^t$ is the advantage function given by:
\begin{equation}
    A^t(T) = A^t(s,a,r,s') = r + V_{\phi_t}(s') - V_{\phi_t}(s)
\end{equation}
where $a \sim \pi_{\theta_t}(\cdot | s)$ and $V_\phi$ is the value network.
$g$ is the clipping function and $\epsilon$ is the clipping factor used by PPO to ensure that updates do not change the policy too much. It is defined as:
\begin{equation}
    g(\epsilon, A) =
        \begin{cases}
            (1 + \epsilon)A & A \geq 0 \\
            (1 - \epsilon)A & A < 0 \\
        \end{cases}
\end{equation}

\section{Experimental Evaluation}
\label{sec:Expt Results}

\subsection{Dataset Details}
To evaluate the efficacy of our proposed method, we train and generate EDA sessions on two types of datasets:
\subsubsection{Cyber Security Datasets}
\begin{wraptable}{r}{4cm}
    \footnotesize
    \centering
    \begin{tabular}{ccc}
    \toprule
    \textbf{Dataset} & \textbf{REACT} & \textbf{Gold} \\
    \midrule
    \textbf{1} & 104 & 7 \\
    \textbf{2} & 79 & 7 \\
    \textbf{3} & 76 & 7 \\
    \textbf{4} & 59 & 7 \\
    \bottomrule
    \end{tabular}
    \caption{Trajectory distribution for training across cyber security datasets}
    \label{tab:dataset_distri}
\end{wraptable}
This is a collection of four mutually exclusive datasets \cite{spitzner2003honeynet}, which we will refer to as the \emph{cyber security datasets}. Expert EDA sessions have been generated in two ways for each of the four datasets: \textbf{(1) REACT Dataset: } 56 cyber security analysts were asked to explore the four cyber security datasets. Each dataset may reveal a unique security event, and analysts were asked to discover the specifics of the underlying security event for each dataset using as many as necessary actions. The REACT dataset is a collection of EDA session traces obtained from these experienced analysts' data exploration on the cyber security datasets and was curated by authors of \cite{milo2018next}.   
\textbf{(2) Gold Dataset: }The gold-standard EDA sessions are generated from walkthrough documents from cyber-security experts, which guide the EDA process for viewers and highlight essential insights in the dataset. Authors curate this from \cite{bar2020automatically}. Table \ref{tab:dataset_distri} shows the number of EDA trajectories found for each of these datasets. We discuss in the following sections how we use these different trajectories for training and testing purposes.



\subsubsection{Synthetic Datasets}
\label{syn-data-method}
 We algorithmically generate datasets with a given schema and inject interesting patterns into them. The injected patterns are in the form of a dependencies between columns which induce a graph structure. We traverse this graph of dependencies to generate \emph{expert} EDA trajectories. Five synthetic datasets, each with three categorical columns, two text columns and three numeric columns and 1000 rows, and their corresponding expert trajectories are generated for our experiments.The generated trajectories are randomly split into train and eval sets. Further details on the generation process are provided in Appendix \ref{appendix:synth-data-gen}.

\subsection{Baselines}
\label{subsec:expt-setup}

\begin{enumerate}
    \item \textbf{Behavioural Cloning}: Behavioural Cloning (BC) is an imitation learning technique that uses expert demonstrations to learn a state-to-action mapping in a supervised way. We use it as a baseline to compare against ILAEDA. For the Cyber Security datasets, we train a BC model on three out of four datasets and report results on sessions generated on the fourth dataset. We train a BC model on the first five tables for the Synthetic datasets and report results for sessions generated on tables $6$ and $7$.
    \item \textbf{ATENA}: We use ATENA \cite{bar2020automatically} as a baseline to compare against our approach. We use the open-source code provided by the authors \cite{Ori2020}. Since ATENA is an RL method that does not use labeled training data, we train and test ATENA models' performance by generating and evaluating trajectories on the same datasets they have been trained on. We train an ATENA model for each Cyber Security dataset and datasets $6$ and $7$ of the synthetic datasets using default parameters from \cite{Ori2020} and report scores by comparing against gold trajectories from each training dataset.  
 \end{enumerate} 

To test the generalization capabilities of ILAEDA, we set up our training and evaluation so that evaluation is performed only on datasets \textbf{not seen at training time}. We train our model for the Cyber Security datasets using a \emph{leave one out} strategy. The model is trained using REACT and gold trajectories from three out of four datasets while keeping the remaining one as a test dataset. This model is then evaluated by generating trajectories on the test dataset and comparing these with the gold-standard trajectories for that dataset. This is repeated four times for each of the four Cyber Security datasets, and we calculate scores to evaluate performance on each left-out dataset. For the synthetic datasets, we train on all the trajectories from the first $5$ of the $7$ datasets. For evaluation, we generate trajectories on the remaining two datasets and report results by comparing them with the evaluation trajectories of the left-out datasets.

\subsection{Evaluation Metrics}
\label{sec:evaluation-metrics}

We evaluate the performance of ILAEDA and the baselines using the following similarity metrics introduced in \cite{bar2020automatically} for evaluation.
\begin{enumerate}
    \item \textbf{Precision:} Precision measures the accuracy of the generated EDA sessions and is calculated as the number of times a view occurs in the gold standard divided by the total number of views. 

    \item \textbf{TBLEU score:} This measure is adopted from BLEU \cite{papineni2002bleu} score.
    It is stricter than precision as it compares subsequences of size n (1 to 3) and considers the order and prevalence of each view in the gold-standard set. The benchmark measures TBLEU-1, TBLEU-2, and TBLEU-3.

    \item \textbf{EDA-Sim:} EDA-Sim \cite{milo2018next}, considers the order of views and enables a fine-grained comparison of EDA views. It considers nearly identical views as \emph{hits}. The generated session is compared to each gold-standard session to get the final EDA-Sim score, using the highest score.
\end{enumerate}

\subsection{Implementation Details}

\subsubsection{Architecture}

ILAEDA's policy and value networks are implemented as fully-connected neural networks with three hidden layers. There are 50 neurons in each layer with $tanh$ non-linearities. The final output layer consists of multiple heads for predicting different components of the action. The design of this layer is identical to that of \cite{bar2020automatically}. The discriminator network is a fully-connected neural network with two hidden layers of 32 neurons each, with ReLU non-linearities.

\subsubsection{Training}

The pre-training phase for ILAEDA is identical to the training for the Behavioural Cloning baseline. It consists of training for 100 epochs with a learning rate of $1e-4$. A batch-size of 32 is used. The adversarial training phase of ILAEDA involves sampling a total of $100,000$ interactions from the environment throughout the training, which corresponds to $\sim 8000$ episodes. A learning rate of $1e-6$ is used for all neural networks. A batch-size of 32 is used for training the policy and value networks, whereas the discriminator is trained with a batch-size of 192. We use a discounting factor of $\gamma = 0.99$. ATENA is trained using default parameters in \cite{Ori2020}.

\section{Results and Analysis}
\label{sec:Analysis}

\begin{table}[htb]
    \centering
    \footnotesize
    \begin{tabular}{lccccc}
    \toprule
     & \textbf{Precision} & \textbf{TBLEU-1} & \textbf{TBLEU-2} & \textbf{TBLEU-3} & \textbf{EDA-sim} \\
    \midrule
    Without BC & 0.259 & 0.103 & 0.038 & 0.029 & 0.195 \\ 
    Without penalty & 0.413 & 0.239 & 0.144 & 0.058 & 0.258 \\ 
    \textbf{ILAEDA (Ours)} & \textbf{0.417} & \textbf{0.271} & \textbf{0.168} & \textbf{0.078} & \textbf{0.310} \\
    \bottomrule
    \end{tabular}
    \caption{Ablation study scores for Cyber Security datasets. The low performance of ablated models validates our choice of penalties and pretraining using BC.}
    \label{tab:ablationscores}
\end{table}

Table \ref{tab:benchmarkscores} shows a performance comparison between ILAEDA and the baselines on the evaluation metrics discussed in Section \ref{sec:evaluation-metrics}. We see that our method overall delivers a significantly better performance than the baselines. The results are all the more significant because, in each setting, ILAEDA is evaluated on datasets that it has not seen during training. In contrast, ATENA is evaluated on the same datasets it trains on. This indicates that ILAEDA generates better EDA sessions and has better generalization capability. 

\subsection{Ablation studies}
\label{section-ablation}

\subsubsection{Training without penalty scores: }
We trained a version of ILAEDA without the coherence penalties defined in \eqref{eq:gail-penalty-reward} and kept all other configurations the same to study the impact of our design choice. We see that this model underperforms in comparison to ILAEDA trained with coherence penalties (Table \ref{tab:ablationscores}). We explain the lower scores by inspecting the generated EDA session (Table \ref{tab:without_penalty_traj}). The EDA views 2 and 3 are identical because the same GROUP operators have been repeated consecutively. Such actions are penalized in our original model. 
\begin{table}[h]
\centering
\footnotesize
\begin{tabular}{l}
\toprule
FILTER eth\_src EQ  00:26:b9:2b:0b:59 \\
\textcolor{red}{GROUP ip\_src AGGREGATE COUNT length} \\
\textcolor{red}{GROUP ip\_src AGGREGATE COUNT length} \\
BACK \\
GROUP tcp\_srcport AGGREGATE SUM ip\_dst \\
BACK \\
GROUP info\_line AGGREGATE SUM eth\_src \\
GROUP tcp\_srcport AGGREGATE SUM length \\
GROUP info\_line AGGREGATE SUM length \\
BACK \\
\bottomrule
\end{tabular}
\caption{\label{tab:without_penalty_traj}EDA session without penalty score on Cyber Security dataset 1. Spurious repetition (red colored) of actions is seen.}
\end{table}

\subsubsection{Training without BC initialization: }
ILAEDA, which is pre-trained with Behavioural Cloning, performs better than a model without pretraining. This result suggests that pretraining the policy before training on the GAIL-based objective improves the overall model. We conjecture that this phenomenon is observed due to the pre-trained policy getting a \textit{warm} start allowing it to start in a better position which gives it an advantage during the adversarial training.

\subsection{Analysis of ILAEDA’s Ability to Capture A Diverse Set of Interestingness Measures}

In Table~\ref{table:traj-ILAEDA}, we show a session generated by our model on Cyber Security dataset 1 and normalized scores obtained on different interestingness metrics. We highlight scores greater than 0.7. Throughout this session, generated views obtained high readability, diversity, and peculiarity scores. This indicates that our model can effectively filter the dataset and present important tuples to users. Views presented are diverse, showing different insights from the dataset.

\begin{table*}[t]
    \centering
    \resizebox{\textwidth}{!}{
    \begin{tabular}{clccccc} 
    \toprule
    \textbf{ Action \#} & \textbf{ ILAEDA Trajectory}                           & \textbf{ A-INT} & \textbf{ Diversity} & \textbf{ Coherence} & \textbf{ Readability} & \textbf{ Peculiarity}  \\ 
    \midrule
    1                   & FILTER eth\_src EQ 00:26:b9:2b:0b:59                  & \textbf{1.00}   & \textbf{0.89}       & \textbf{0.83}       & \textbf{1.00}         & 0.00                   \\
    2                   & FILTER ip\_src NEQ 82.108.69.238                      & 0.05            & 0.62                & \textbf{0.90}       & \textbf{0.95}         & 0.00                   \\
    3                   & FILTER highest\_layer NEQ ICMP                        & 0.49            & \textbf{0.73}       & 0.00                & \textbf{0.99}         & \textbf{0.97}          \\
    4                   & FILTER highest\_layer NEQ ARP                         & 0.05            & 0.68                & 0.00                & \textbf{0.95}         & \textbf{0.97}          \\
    5                   & FILTER captured\_length NEQ 62                        & 0.05            & 0.68                & 0.00                & \textbf{0.95}         & \textbf{1.00}          \\
    6                   & GROUP captured\_length AGGREGATE COUNT packet\_number & 0.00            & \textbf{1.00}       & 0.00                & 0.00                  & \textbf{1.00}          \\
    7                   & BACK                                                  & 0.00            & 0.00                & \textbf{0.80}       & \textbf{0.85}         & \textbf{1.00}          \\
    8                   & GROUP highest\_layer AGGREGATE COUNT length           & 0.00            & \textbf{0.81}       & 0.00                & 0.00                  & \textbf{1.00}          \\
    9                   & GROUP ip\_src AGGREGATE COUNT length                  & 0.14            & \textbf{0.92}       & \textbf{1.00}       & \textbf{0.85}         & \textbf{1.00}          \\
    10                  & GROUP tcp\_srcport AGGREGATE COUNT length             & 0.07            & \textbf{0.77}       & \textbf{0.88}       & \textbf{0.92}         & \textbf{1.00}          \\
    \bottomrule
    \end{tabular}
    }
    \caption{Trajectory generated by the ILAEDA for Cyber Security dataset 1 and normalized values attained by different interestingness measures for each action.}
    \label{table:traj-ILAEDA}
\end{table*}

\begin{table*}[htb]
    \centering
    \resizebox{\textwidth}{!}{
    \begin{tabular}{clccccc} 
    \toprule
    \textbf{ Action \#} & \textbf{ ATENA Trajectory}                          & \textbf{ A-INT} & \textbf{ Diversity} & \textbf{ Coherence} & \textbf{ Readability} & \textbf{ Peculiarity}  \\ 
    \midrule
    1                   & GROUP highest\_layer AGGREGATE COUNT packet\_number & \textbf{1.00}   & \textbf{0.90}       & \textbf{1.00}       & \textbf{1.00}         & \textbf{0.71}          \\
    2                   & GROUP eth\_src AGGREGATE COUNT packet\_number       & \textbf{0.89}   & \textbf{0.72}       & 0.53                & 0.10                  & \textbf{1.00}          \\
    3                   & BACK                                                & 0.00            & 0.00                & 0.67                & 0.10                  & \textbf{0.71}          \\
    4                   & BACK                                                & 0.00            & 0.00                & \textbf{0.80}       & 0.09                  & 0.00                   \\
    5                   & GROUP eth\_dst AGGREGATE COUNT packet\_number       & \textbf{1.00}   & \textbf{0.72}       & 0.53                & \textbf{1.00}         & 0.61                   \\
    6                   & FILTER tcp\_srcport CONTAINS 139.0                  & \textbf{0.79}   & \textbf{0.96}       & 0.00                & 0.10                  & 0.08                   \\
    7                   & GROUP length AGGREGATE COUNT packet\_number         & \textbf{0.94}   & 0.66                & 0.53                & 0.10                  & 0.11                   \\
    8                   & GROUP eth\_src AGGREGATE COUNT packet\_number       & \textbf{0.93}   & 0.60                & 0.51                & 0.10                  & 0.11                   \\
    9                   & GROUP highest\_layer AGGREGATE COUNT packet\_number & \textbf{0.91}   & 0.59                & 0.54                & 0.10                  & 0.11                   \\
    10                  & GROUP ip\_dst AGGREGATE COUNT packet\_number        & \textbf{0.89}   & 0.58                & 0.00                & 0.10                  & 0.11                   \\
    11                  & GROUP tcp\_srcport AGGREGATE COUNT packet\_number   & \textbf{0.87}   & 0.58                & 0.44                & 0.10                  & 0.11                   \\
    12                  & FILTER ip\_src CONTAINS 82.108.10.135               & 0.46            & \textbf{1.00}       & 0.00                & 0.00                  & 0.00                   \\
    \bottomrule
    \end{tabular}
    }
    \caption{Trajectory generated by ATENA model for Cyber Security Dataset 1 and normalized values attained by different interestingness measures for each action.}
    \label{table:traj-ATENA}
\end{table*}
Table~\ref{table:traj-ATENA} shows a session generated using ATENA on the same dataset. We observe that views produced have high A-INT and diversity scores. This is because these scores were used as reward signals to train ATENA. In comparison, each action taken by our model shows different scores being highlighted, which shows that our model can capture a wider set of interestingness measures than ATENA.

\subsection{Analysis of ILAEDA's Ability to Capture Specific Interestingness Measures}

\begin{figure*}[htb]
  \centering
  \includegraphics[width=0.5\textwidth]{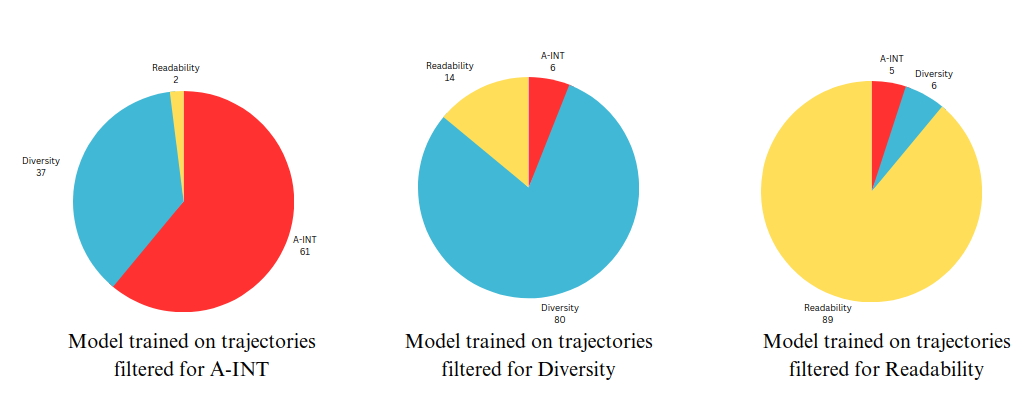}
  \caption{Distribution of metrics in sessions generated on Synthetic Dataset 7 using ILAEDA models. ILAEDA models were trained on sessions filtered to maximize A-INT, Diversity, or Readability more than other metrics. }
  \label{fig:trajectory-distribution}
\end{figure*}

To analyze ILAEDA's ability to capture specific underlying interestingness measures 
we conducted an experiment where we trained ILAEDA on a subset of expert sessions
filtered to optimize one specific measure on a source datasets. The goal is to examine 
if the sessions generated by the trained model on other datasets also reflect the domination of the same interestingness metric.

We filter the expert sessions from the Synthetic Datasets (refer to section \ref{syn-data-method}) to obtain subsets where the sessions maximize one particular interestingness metric over others. We consider A-INT, Diversity, and Readability as the metrics for this analysis. To filter the expert sessions, we classify each session into one of three categories. (1) {\bf Category 1: }Maximizing A-INT more than others. (2) {\bf Category 2: }Maximizing Diversity more than others. (3) {\bf Category 3: }Maximizing Readability more than others.

We compare the $75^{th}$ percentile normalized score obtained for each metric over all the steps in that session's trajectory to classify a session. If the $75^{th}$ percentile normalized score over that trajectory is highest for the A-INT metric, then that session is classified as maximizing A-INT more than others. Similarly, we can classify a session as maximizing Diversity more than others or maximizing Readability more than others.

We do this classification for all Synthetic datasets 1-5 sessions and obtain three subsets. We then trained an ILAEDA model on these subsets using the default settings described before. We use the trained models to generate 100 EDA sessions each on Synthetic Dataset 7. For each generated session, we compare the median normalized score obtained for each metric across all the steps in the trajectory of that session. If the median normalized score is highest for A-INT, we say this session maximizes A-INT.
Similarly, a session can also be classified as maximizing Readability/Diversity more than other metrics. Figure \ref{fig:trajectory-distribution} shows the distribution of most-maximized metrics for sessions generated using each of the three ILAEDA models. We see that in all three cases, most sessions maximize that interestingness metric more than others, for which the training sessions used to train the model generating those sessions were filtered. A model trained on sessions filtered for maximizing A-INT more than others tends to maximize A-INT more than other metrics when tested on an unseen dataset. This analysis strongly suggests that our method can capture specific underlying interestingness measures being optimized for in a set of expert demonstrations, and the trained model can also optimize for those measures on unseen datasets.

\section{Conclusion and Future Work}
In this paper, we present ILAEDA, a novel imitation learning based end-to-end AutoEDA system which does not rely on handcrafted interestingness heuristics to learn a policy. Through automated benchmarking and manual analysis of generated EDA sessions, we demonstrate that ILAEDA can learn to perform meaningful EDA and outperform baselines across benchmarks. Further, we demonstrate the generalizability of our system on unseen datasets. We also show ablation studies to validate our design choices. 
For future work, some of the following investigations may be worthwhile: (a) training a system that generalizes to perform AutoEDA across datasets with different schemas, (b) exploring the effect of other penalties and (c) incorporating actions other than FILTER and GROUP into ILAEDA.

\bibliographystyle{ACM-Reference-Format}
\bibliography{sample-base}


\begin{thebibliography}{37}


\ifx \showCODEN    \undefined \def \showCODEN     #1{\unskip}     \fi
\ifx \showDOI      \undefined \def \showDOI       #1{#1}\fi
\ifx \showISBNx    \undefined \def \showISBNx     #1{\unskip}     \fi
\ifx \showISBNxiii \undefined \def \showISBNxiii  #1{\unskip}     \fi
\ifx \showISSN     \undefined \def \showISSN      #1{\unskip}     \fi
\ifx \showLCCN     \undefined \def \showLCCN      #1{\unskip}     \fi
\ifx \shownote     \undefined \def \shownote      #1{#1}          \fi
\ifx \showarticletitle \undefined \def \showarticletitle #1{#1}   \fi
\ifx \showURL      \undefined \def \showURL       {\relax}        \fi
\providecommand\bibfield[2]{#2}
\providecommand\bibinfo[2]{#2}
\providecommand\natexlab[1]{#1}
\providecommand\showeprint[2][]{arXiv:#2}

\bibitem[Aligon et~al\mbox{.}(2015)]%
        {aligon2015collaborative}
\bibfield{author}{\bibinfo{person}{Julien Aligon}, \bibinfo{person}{Enrico Gallinucci}, \bibinfo{person}{Matteo Golfarelli}, \bibinfo{person}{Patrick Marcel}, {and} \bibinfo{person}{Stefano Rizzi}.} \bibinfo{year}{2015}\natexlab{}.
\newblock \showarticletitle{A collaborative filtering approach for recommending OLAP sessions}.
\newblock \bibinfo{journal}{\emph{Decision Support Systems}}  \bibinfo{volume}{69} (\bibinfo{year}{2015}), \bibinfo{pages}{20--30}.
\newblock


\bibitem[Bar~El et~al\mbox{.}(2020a)]%
        {Ori2020}
\bibfield{author}{\bibinfo{person}{Ori Bar~El}, \bibinfo{person}{Tova Milo}, {and} \bibinfo{person}{Amit Somech}.} \bibinfo{year}{2020}\natexlab{a}.
\newblock \bibinfo{title}{ATENA Basic Implementation}.
\newblock \bibinfo{howpublished}{\url{https://github.com/TAU-DB/ATENA-A-EDA/tree/master/atena-basic}}.
\newblock


\bibitem[Bar~El et~al\mbox{.}(2020b)]%
        {bar2020automatically}
\bibfield{author}{\bibinfo{person}{Ori Bar~El}, \bibinfo{person}{Tova Milo}, {and} \bibinfo{person}{Amit Somech}.} \bibinfo{year}{2020}\natexlab{b}.
\newblock \showarticletitle{Automatically generating data exploration sessions using deep reinforcement learning}. In \bibinfo{booktitle}{\emph{ACM SIGMOD}}. \bibinfo{pages}{1527--1537}.
\newblock


\bibitem[Cao et~al\mbox{.}(2022)]%
        {Cao2022LearnTE}
\bibfield{author}{\bibinfo{person}{Yukun Cao}, \bibinfo{person}{Xike Xie}, {and} \bibinfo{person}{Kexin Huang}.} \bibinfo{year}{2022}\natexlab{}.
\newblock \showarticletitle{Learn to Explore: on Bootstrapping Interactive Data Exploration with Meta-learning}.
\newblock \bibinfo{journal}{\emph{IEEE ICDE}} (\bibinfo{year}{2022}), \bibinfo{pages}{1720--1733}.
\newblock


\bibitem[Chandola and Kumar(2007)]%
        {chandola2007summarization}
\bibfield{author}{\bibinfo{person}{Varun Chandola} {and} \bibinfo{person}{Vipin Kumar}.} \bibinfo{year}{2007}\natexlab{}.
\newblock \showarticletitle{Summarization--compressing data into an informative representation}.
\newblock \bibinfo{journal}{\emph{Knowledge and Information Systems}} \bibinfo{volume}{12}, \bibinfo{number}{3} (\bibinfo{year}{2007}), \bibinfo{pages}{355--378}.
\newblock


\bibitem[Chanson(2022)]%
        {Chanson2022AutomaticGO}
\bibfield{author}{\bibinfo{person}{Alexandre Chanson}.} \bibinfo{year}{2022}\natexlab{}.
\newblock \showarticletitle{Automatic generation of comparison notebooks for interactive data exploration}. In \bibinfo{booktitle}{\emph{EDBT}}.
\newblock


\bibitem[Ding et~al\mbox{.}(2019)]%
        {NEURIPS2019_c8d3a760}
\bibfield{author}{\bibinfo{person}{Yiming Ding}, \bibinfo{person}{Carlos Florensa}, \bibinfo{person}{Pieter Abbeel}, {and} \bibinfo{person}{Mariano Phielipp}.} \bibinfo{year}{2019}\natexlab{}.
\newblock \showarticletitle{Goal-conditioned Imitation Learning}. In \bibinfo{booktitle}{\emph{NeurIPS}}, Vol.~\bibinfo{volume}{32}.
\newblock


\bibitem[Drosou and Pitoura(2013)]%
        {drosou2013ymaldb}
\bibfield{author}{\bibinfo{person}{Marina Drosou} {and} \bibinfo{person}{Evaggelia Pitoura}.} \bibinfo{year}{2013}\natexlab{}.
\newblock \showarticletitle{Ymaldb: exploring relational databases via result-driven recommendations}.
\newblock \bibinfo{journal}{\emph{The VLDB Journal}} \bibinfo{volume}{22}, \bibinfo{number}{6} (\bibinfo{year}{2013}), \bibinfo{pages}{849--874}.
\newblock


\bibitem[Eirinaki et~al\mbox{.}(2013)]%
        {eirinaki2013querie}
\bibfield{author}{\bibinfo{person}{Magdalini Eirinaki}, \bibinfo{person}{Suju Abraham}, \bibinfo{person}{Neoklis Polyzotis}, {and} \bibinfo{person}{Naushin Shaikh}.} \bibinfo{year}{2013}\natexlab{}.
\newblock \showarticletitle{Querie: Collaborative database exploration}.
\newblock \bibinfo{journal}{\emph{IEEE TKDE}} \bibinfo{volume}{26}, \bibinfo{number}{7} (\bibinfo{year}{2013}), \bibinfo{pages}{1778--1790}.
\newblock


\bibitem[Garg et~al\mbox{.}(2021)]%
        {Garg2021IQLearnIS}
\bibfield{author}{\bibinfo{person}{Divyansh Garg}, \bibinfo{person}{Shuvam Chakraborty}, \bibinfo{person}{Chris Cundy}, \bibinfo{person}{Jiaming Song}, {and} \bibinfo{person}{Stefano Ermon}.} \bibinfo{year}{2021}\natexlab{}.
\newblock \showarticletitle{IQ-Learn: Inverse soft-Q Learning for Imitation}.
\newblock \bibinfo{journal}{\emph{ArXiv}}  \bibinfo{volume}{abs/2106.12142} (\bibinfo{year}{2021}).
\newblock


\bibitem[Geng and Hamilton(2006)]%
        {geng2006interestingness}
\bibfield{author}{\bibinfo{person}{Liqiang Geng} {and} \bibinfo{person}{Howard~J Hamilton}.} \bibinfo{year}{2006}\natexlab{}.
\newblock \showarticletitle{Interestingness measures for data mining: A survey}.
\newblock \bibinfo{journal}{\emph{ACM Computing Surveys (CSUR)}} \bibinfo{volume}{38}, \bibinfo{number}{3} (\bibinfo{year}{2006}), \bibinfo{pages}{9--es}.
\newblock


\bibitem[Ho and Ermon(2016)]%
        {ho2016generative}
\bibfield{author}{\bibinfo{person}{Jonathan Ho} {and} \bibinfo{person}{Stefano Ermon}.} \bibinfo{year}{2016}\natexlab{}.
\newblock \showarticletitle{Generative adversarial imitation learning}.
\newblock \bibinfo{journal}{\emph{NIPS}}  \bibinfo{volume}{29} (\bibinfo{year}{2016}).
\newblock


\bibitem[Joglekar et~al\mbox{.}(2017)]%
        {joglekar2017interactive}
\bibfield{author}{\bibinfo{person}{Manas Joglekar}, \bibinfo{person}{Hector Garcia-Molina}, {and} \bibinfo{person}{Aditya Parameswaran}.} \bibinfo{year}{2017}\natexlab{}.
\newblock \showarticletitle{Interactive data exploration with smart drill-down}.
\newblock \bibinfo{journal}{\emph{IEEE TKDE}} \bibinfo{volume}{31}, \bibinfo{number}{1} (\bibinfo{year}{2017}), \bibinfo{pages}{46--60}.
\newblock


\bibitem[Kullback and Leibler(1951)]%
        {kullback1951information}
\bibfield{author}{\bibinfo{person}{Solomon Kullback} {and} \bibinfo{person}{Richard~A Leibler}.} \bibinfo{year}{1951}\natexlab{}.
\newblock \showarticletitle{On information and sufficiency}.
\newblock \bibinfo{journal}{\emph{The annals of mathematical statistics}} \bibinfo{volume}{22}, \bibinfo{number}{1} (\bibinfo{year}{1951}), \bibinfo{pages}{79--86}.
\newblock


\bibitem[Lipman et~al\mbox{.}(2023)]%
        {Lipman2023ATENAPROGP}
\bibfield{author}{\bibinfo{person}{Tavor Lipman}, \bibinfo{person}{Tova Milo}, {and} \bibinfo{person}{Amit Somech}.} \bibinfo{year}{2023}\natexlab{}.
\newblock \showarticletitle{ATENA-PRO: Generating Personalized Exploration Notebooks with Constrained Reinforcement Learning}.
\newblock \bibinfo{journal}{\emph{Companion of the 2023 International Conference on Management of Data}} (\bibinfo{year}{2023}).
\newblock


\bibitem[Ma et~al\mbox{.}(2021)]%
        {Ma2021MetaInsightAD}
\bibfield{author}{\bibinfo{person}{Ping Ma}, \bibinfo{person}{Rui Ding}, \bibinfo{person}{Shi Han}, {and} \bibinfo{person}{Dongmei Zhang}.} \bibinfo{year}{2021}\natexlab{}.
\newblock \showarticletitle{MetaInsight: Automatic Discovery of Structured Knowledge for Exploratory Data Analysis}.
\newblock \bibinfo{journal}{\emph{Proceedings of the 2021 International Conference on Management of Data}} (\bibinfo{year}{2021}).
\newblock


\bibitem[Ma et~al\mbox{.}(2023)]%
        {Ma2023InsightPilotAL}
\bibfield{author}{\bibinfo{person}{Pingchuan Ma}, \bibinfo{person}{Rui Ding}, \bibinfo{person}{Shuai Wang}, \bibinfo{person}{Shi Han}, {and} \bibinfo{person}{Dongmei Zhang}.} \bibinfo{year}{2023}\natexlab{}.
\newblock \showarticletitle{InsightPilot: An LLM-Empowered Automated Data Exploration System}. In \bibinfo{booktitle}{\emph{EMNLP}}.
\newblock


\bibitem[Milo and Somech(2016)]%
        {milo2016react}
\bibfield{author}{\bibinfo{person}{Tova Milo} {and} \bibinfo{person}{Amit Somech}.} \bibinfo{year}{2016}\natexlab{}.
\newblock \showarticletitle{React: Context-sensitive recommendations for data analysis}. In \bibinfo{booktitle}{\emph{Proceedings of the 2016 International Conference on Management of Data}}. \bibinfo{pages}{2137--2140}.
\newblock


\bibitem[Milo and Somech(2018)]%
        {milo2018next}
\bibfield{author}{\bibinfo{person}{Tova Milo} {and} \bibinfo{person}{Amit Somech}.} \bibinfo{year}{2018}\natexlab{}.
\newblock \showarticletitle{Next-step suggestions for modern interactive data analysis platforms}. In \bibinfo{booktitle}{\emph{KDD}}. \bibinfo{pages}{576--585}.
\newblock


\bibitem[Papineni et~al\mbox{.}(2002)]%
        {papineni2002bleu}
\bibfield{author}{\bibinfo{person}{Kishore Papineni}, \bibinfo{person}{Salim Roukos}, \bibinfo{person}{Todd Ward}, {and} \bibinfo{person}{Wei-Jing Zhu}.} \bibinfo{year}{2002}\natexlab{}.
\newblock \showarticletitle{Bleu: a method for automatic evaluation of machine translation}. In \bibinfo{booktitle}{\emph{ACL}}. \bibinfo{pages}{311--318}.
\newblock


\bibitem[Personnaz et~al\mbox{.}(2021a)]%
        {personnaz2021balancing}
\bibfield{author}{\bibinfo{person}{Aur{\'e}lien Personnaz}, \bibinfo{person}{Sihem Amer-Yahia}, \bibinfo{person}{Laure Berti-Equille}, \bibinfo{person}{Maximilian Fabricius}, {and} \bibinfo{person}{Srividya Subramanian}.} \bibinfo{year}{2021}\natexlab{a}.
\newblock \showarticletitle{Balancing Familiarity and Curiosity in Data Exploration with Deep Reinforcement Learning}. In \bibinfo{booktitle}{\emph{Workshop in Exploiting AI Techniques for Data Management}}. \bibinfo{pages}{16--23}.
\newblock


\bibitem[Personnaz et~al\mbox{.}(2021b)]%
        {Personnaz2021DORATE}
\bibfield{author}{\bibinfo{person}{Aur{\'e}lien Personnaz}, \bibinfo{person}{Sihem Amer-Yahia}, \bibinfo{person}{Laure Berti-{\'E}quille}, \bibinfo{person}{Maximilian Fabricius}, {and} \bibinfo{person}{Srividya Subramanian}.} \bibinfo{year}{2021}\natexlab{b}.
\newblock \showarticletitle{DORA THE EXPLORER: Exploring Very Large Data With Interactive Deep Reinforcement Learning}.
\newblock \bibinfo{journal}{\emph{ACM CIKM}} (\bibinfo{year}{2021}).
\newblock


\bibitem[Razmadze et~al\mbox{.}(2022)]%
        {Razmadze2022SelectingSF}
\bibfield{author}{\bibinfo{person}{Kathy Razmadze}, \bibinfo{person}{Yael Amsterdamer}, \bibinfo{person}{Amit Somech}, \bibinfo{person}{Susan~B. Davidson}, {and} \bibinfo{person}{Tova Milo}.} \bibinfo{year}{2022}\natexlab{}.
\newblock \showarticletitle{Selecting Sub-tables for Data Exploration}.
\newblock \bibinfo{journal}{\emph{IEEE ICDE}} (\bibinfo{year}{2022}), \bibinfo{pages}{2496--2509}.
\newblock


\bibitem[Reddy et~al\mbox{.}(2019)]%
        {Reddy2019SQILIL}
\bibfield{author}{\bibinfo{person}{Siddharth Reddy}, \bibinfo{person}{Anca~D. Dragan}, {and} \bibinfo{person}{Sergey Levine}.} \bibinfo{year}{2019}\natexlab{}.
\newblock \showarticletitle{SQIL: Imitation Learning via Reinforcement Learning with Sparse Rewards}.
\newblock \bibinfo{journal}{\emph{arXiv: Learning}} (\bibinfo{year}{2019}).
\newblock
\urldef\tempurl%
\url{https://api.semanticscholar.org/CorpusID:202888699}
\showURL{%
\tempurl}


\bibitem[Ross and Bagnell(2010)]%
        {Ross2010EfficientRF}
\bibfield{author}{\bibinfo{person}{St{\'e}phane Ross} {and} \bibinfo{person}{Drew Bagnell}.} \bibinfo{year}{2010}\natexlab{}.
\newblock \showarticletitle{Efficient Reductions for Imitation Learning}. In \bibinfo{booktitle}{\emph{AISTATS}}.
\newblock


\bibitem[Sarawagi(2000)]%
        {sarawagi2000user}
\bibfield{author}{\bibinfo{person}{Sunita Sarawagi}.} \bibinfo{year}{2000}\natexlab{}.
\newblock \showarticletitle{User-adaptive exploration of multidimensional data}. In \bibinfo{booktitle}{\emph{VLDB}}, Vol.~\bibinfo{volume}{2000}. Citeseer, \bibinfo{pages}{307--316}.
\newblock


\bibitem[Sarawagi et~al\mbox{.}(1998)]%
        {sarawagi1998discovery}
\bibfield{author}{\bibinfo{person}{Sunita Sarawagi}, \bibinfo{person}{Rakesh Agrawal}, {and} \bibinfo{person}{Nimrod Megiddo}.} \bibinfo{year}{1998}\natexlab{}.
\newblock \showarticletitle{Discovery-driven exploration of OLAP data cubes}. In \bibinfo{booktitle}{\emph{International Conference on Extending Database Technology}}. Springer, \bibinfo{pages}{168--182}.
\newblock


\bibitem[Schulman et~al\mbox{.}(2017)]%
        {schulman2017proximal}
\bibfield{author}{\bibinfo{person}{John Schulman}, \bibinfo{person}{Filip Wolski}, \bibinfo{person}{Prafulla Dhariwal}, \bibinfo{person}{Alec Radford}, {and} \bibinfo{person}{Oleg Klimov}.} \bibinfo{year}{2017}\natexlab{}.
\newblock \showarticletitle{Proximal policy optimization algorithms}.
\newblock \bibinfo{journal}{\emph{arXiv preprint arXiv:1707.06347}} (\bibinfo{year}{2017}).
\newblock


\bibitem[Singh et~al\mbox{.}(2016)]%
        {singh2016dbexplorer}
\bibfield{author}{\bibinfo{person}{Manish Singh}, \bibinfo{person}{Michael~J Cafarella}, {and} \bibinfo{person}{HV Jagadish}.} \bibinfo{year}{2016}\natexlab{}.
\newblock \showarticletitle{DBExplorer: Exploratory Search in Databases.}. In \bibinfo{booktitle}{\emph{EDBT}}. \bibinfo{pages}{89--100}.
\newblock


\bibitem[Somech et~al\mbox{.}(2019)]%
        {somech2019predicting}
\bibfield{author}{\bibinfo{person}{Amit Somech}, \bibinfo{person}{Tova Milo}, {and} \bibinfo{person}{Chai Ozeri}.} \bibinfo{year}{2019}\natexlab{}.
\newblock \showarticletitle{Predicting" what is interesting" by mining interactive-data-analysis session logs}. In \bibinfo{booktitle}{\emph{EDBT}}.
\newblock


\bibitem[Spitzner(2003)]%
        {spitzner2003honeynet}
\bibfield{author}{\bibinfo{person}{Lance Spitzner}.} \bibinfo{year}{2003}\natexlab{}.
\newblock \showarticletitle{The honeynet project: Trapping the hackers}.
\newblock \bibinfo{journal}{\emph{IEEE Security \& Privacy}} \bibinfo{volume}{1}, \bibinfo{number}{2} (\bibinfo{year}{2003}), \bibinfo{pages}{15--23}.
\newblock


\bibitem[Torabi et~al\mbox{.}(2018)]%
        {10.5555/3304652.3304697}
\bibfield{author}{\bibinfo{person}{Faraz Torabi}, \bibinfo{person}{Garrett Warnell}, {and} \bibinfo{person}{Peter Stone}.} \bibinfo{year}{2018}\natexlab{}.
\newblock \showarticletitle{Behavioral Cloning from Observation}. In \bibinfo{booktitle}{\emph{IJCAI}}. \bibinfo{pages}{4950–4957}.
\newblock


\bibitem[Touvron et~al\mbox{.}(2023)]%
        {touvron2023llama}
\bibfield{author}{\bibinfo{person}{Hugo Touvron}, \bibinfo{person}{Thibaut Lavril}, \bibinfo{person}{Gautier Izacard}, \bibinfo{person}{Xavier Martinet}, \bibinfo{person}{Marie-Anne Lachaux}, \bibinfo{person}{Timothée Lacroix}, \bibinfo{person}{Baptiste Rozière}, \bibinfo{person}{Naman Goyal}, \bibinfo{person}{Eric Hambro}, \bibinfo{person}{Faisal Azhar}, \bibinfo{person}{Aurelien Rodriguez}, \bibinfo{person}{Armand Joulin}, \bibinfo{person}{Edouard Grave}, {and} \bibinfo{person}{Guillaume Lample}.} \bibinfo{year}{2023}\natexlab{}.
\newblock \bibinfo{title}{LLaMA: Open and Efficient Foundation Language Models}.
\newblock
\newblock
\showeprint[arxiv]{2302.13971}~[cs.CL]


\bibitem[van Leeuwen(2010)]%
        {van2010maximal}
\bibfield{author}{\bibinfo{person}{Matthijs van Leeuwen}.} \bibinfo{year}{2010}\natexlab{}.
\newblock \showarticletitle{Maximal exceptions with minimal descriptions}.
\newblock \bibinfo{journal}{\emph{DMKD}} \bibinfo{volume}{21}, \bibinfo{number}{2} (\bibinfo{year}{2010}), \bibinfo{pages}{259--276}.
\newblock


\bibitem[Vartak et~al\mbox{.}(2015)]%
        {vartak2015seedb}
\bibfield{author}{\bibinfo{person}{Manasi Vartak}, \bibinfo{person}{Sajjadur Rahman}, \bibinfo{person}{Samuel Madden}, \bibinfo{person}{Aditya Parameswaran}, {and} \bibinfo{person}{Neoklis Polyzotis}.} \bibinfo{year}{2015}\natexlab{}.
\newblock \showarticletitle{Seedb: Efficient data-driven visualization recommendations to support visual analytics}. In \bibinfo{booktitle}{\emph{VLDB}}, Vol.~\bibinfo{volume}{8}. NIH Public Access, \bibinfo{pages}{2182}.
\newblock


\bibitem[Yang et~al\mbox{.}(2009)]%
        {yang2009recommending}
\bibfield{author}{\bibinfo{person}{Xiaoyan Yang}, \bibinfo{person}{Cecilia~M Procopiuc}, {and} \bibinfo{person}{Divesh Srivastava}.} \bibinfo{year}{2009}\natexlab{}.
\newblock \showarticletitle{Recommending join queries via query log analysis}. In \bibinfo{booktitle}{\emph{IEEE ICDE}}. IEEE, \bibinfo{pages}{964--975}.
\newblock


\bibitem[Ziebart et~al\mbox{.}(2008)]%
        {10.5555/1620270.1620297}
\bibfield{author}{\bibinfo{person}{Brian~D. Ziebart}, \bibinfo{person}{Andrew Maas}, \bibinfo{person}{J.~Andrew Bagnell}, {and} \bibinfo{person}{Anind~K. Dey}.} \bibinfo{year}{2008}\natexlab{}.
\newblock \showarticletitle{Maximum Entropy Inverse Reinforcement Learning}. In \bibinfo{booktitle}{\emph{AAAI}}.
\newblock


\end{thebibliography}

\appendix

\section{Interestingness Measures}
\label{appendix:interestingness-measures}

\begin{enumerate}[leftmargin=*]
    \item \textbf{A-INT.} This interestingness score \cite{bar2020automatically} is used as a reward function. It is defined for FILTER and GROUP operations as follows:
    \begin{enumerate}
        \item Interestingness Score for a GROUP operation:  This score uses \textit{conciseness measures} \cite{chandola2007summarization} \cite{geng2006interestingness} which rewards compact group-by views covering many rows as these views are considered informative and easy to understand. This measure considers the number of groups $g$, the number of grouped attributes $a$, and the number of tuples $r$. The score $Int(d_t)$ is calculated as 
        \begin{equation}
            Int(d_t) = \frac{h_1(g \cdot a)}{h_2(r)}
        \end{equation} 
        where $h_1$ and $h_2$ are normalized sigmoid function with fixed width and center. 
        \item Interestingness Score for a FILTER operation: To quantify the interestingness of a FILTER operation, \textit{exceptionality} \cite{sarawagi1998discovery} \cite{van2010maximal} \cite{vartak2015seedb} of filtered rows (generated after applying the operation) is compared with those of the unfiltered table. The Kullback-Leibler (KL) divergence \cite{kullback1951information} on each column is used to measure how much the filtered data view differs from the unfiltered view. This score favors filter operations whose resultant display deviates significantly from the previous display. Let $P_{t}^A$ be the distribution of an attribute (column) $A$ at time step $t$ in the analysis. The score $Int(d_t)$ is given by:
        \begin{equation}
            Int(d_t) = h(\max_{A} D_{KL}(P_{t-1}^A, P_{t}^A))
        \end{equation}
        where $h$ is the sigmoid function.
    \end{enumerate}
    
    \item \textbf{Diversity score.} This metric favors a display that highlights parts of the dataset that are different from those seen in any of the previous displays in the session so far \cite{bar2020automatically}. It is calculated as the minimum Euclidean distance between the 
    resultant display of an operation and 
all previous displays, i.e. 
    \begin{equation}
         Div(d_t) = \min_{0 \leq t' < t} \delta(d_t, d_{t’})
    \end{equation}
    \item \textbf{Readability score.} This metric uses compaction gain from \cite{chandola2007summarization}. This score measures changes in the compactness of the display before and after performing the current EDA operation. Highly compact displays with few rows are considered more readable and are given a higher reward. Let compact display score, $C(d_t)$, for current display $d_t$ with number of groups $g$ be defined as 
    \begin{equation}
        C(d_t) = h_1(g \cdot |d_t|)
    \end{equation}
    where $h_1$ is normalized sigmoid function with fixed width and center. Readability score, $Read(d_t)$ is defined
    \begin{equation}
        Read(d_t) = 1 - \frac{C(d_{t-1})}{C(d_t)}
    \end{equation}

    \item \textbf{Peculiarity score.} This score helps quantify anomalous patterns. This a deviation-based measure from \cite{vartak2015seedb}, which favors display with a high difference from the initial display at the start of the analysis. It is defined in terms of the KL-divergence as:
    \begin{equation}
        Pec(d_t) = h(\max_{A} D_{KL}(P_{0}^A, P_{t}^A))
    \end{equation}

    \item \textbf{Coherence score.} This score is a highly detailed metric determined by a set of handcrafted rules which assign each display a penalty or reward based on whether the operation performed at the current step is coherent with previous operations. There are two types of rules: (1) Rules that apply to any generic EDA session. For example, an operation that results in an empty or unchanged display is considered incoherent. (2) Rules specific to the dataset's domain to be explored. \cite{bar2020automatically} define many such rules for the Cyber security datasets \cite{spitzner2003honeynet} being analyzed. These rules are highly specific and are carefully hand-tuned to predict coherence for actions taken on the Cyber security dataset.
\end{enumerate}

\section{Algorithmic Details of ILAEDA}
\label{sec:ilaeda-algo}

Algorithm \ref{alg:imitalg} gives a detailed description of the ILAEDA procedure.

\begin{algorithm*}
\small
\caption{\bf ILAEDA}\label{alg:imitalg}
\textbf{Section 1:} Pre-training with Behavioral Cloning
\begin{algorithmic}[1]
\STATE \textbf{Input:} Expert trajectories $\mathcal{E}$, initial policy parameters $\theta_0$, number of steps $BC$, $\alpha$
\FOR{i=$0,1,2,\dots, BC$}
    \STATE Sample $\mathcal{T}_E = \{\tau_1, \tau_2, \dots \} \sim \mathcal{E}$
    \STATE Update $\theta_i$ to $\theta_{i+1}$ by
    \begin{equation*}
        \theta_{i+1} = \theta_i + \alpha \nabla_{\theta_i} \left[ \frac{1}{|\mathcal{T}_E|} \frac{1}{T} \sum_{\tau \in \mathcal{T}_E} \sum_{t=0}^{T-1} -\log(\pi_{\theta_i}(a_t | s_t) + || \theta_i ||_2 \right]
    \end{equation*}
\ENDFOR
\STATE \textbf{Output:} Pre-trained policy parameters $\theta_{BC}$
\end{algorithmic}
\textbf{Section 2:} Adversarial Training
\begin{algorithmic}[1]
\STATE \textbf{Input:} Environment, Max Episode length $T$, Expert trajectories $\mathcal{E}$, Pretrained policy parameters $\theta_0 = \theta_{BC}$, randomly initiated value and discriminator parameters $\phi_0$ and $w_0$ respectively, train interval $T_{train}$, learning rate $\alpha$, clipping parameter $\epsilon$ 
\STATE Initialize buffer $\mathcal{B}$ for generated trajectories. 
\FOR{i=1,2,...}
    \STATE Sample trajectories $\mathcal{T}_i = \{ \tau^{\theta_i}_1, \tau^{\theta_i}_2, \dots \}$ by interacting with the environment using $\pi_{\theta_i}$, where each $\tau^{\theta_i} = \{ (s_1,a_1), \dots \{s_T, a_T \}$ 
    \STATE Obtain $(s_t,a_t,r_t,s_{t+1})$ tuples from $\mathcal{T}_i$ with $r_t = - \log (1 - D_{w_i}(s_t, a_t)) + P(s_{1:t}, a_{1:t})$. Add these tuples to $\mathcal{B}$
    \IF{$i = 0 \mod T_{train}$}
        \STATE Sample generated trajectories $\mathcal{T}_G \sim \mathcal{B}$ and expert trajectories $\mathcal{T}_E \sim \mathcal{E}$ such that $|\mathcal{T}_E| = |\mathcal{T}_G|$
        \STATE Update discriminator parameters $w_i$ to $w_{i+1}$ as  \begin{align*}
            \quad\quad\quad w_{i+1} = w_i + \alpha \nabla_{w_i} \left[ \frac{1}{|\mathcal{T}_G|} \frac{1}{T} \sum_{\tau \in \mathcal{T}_G}{\sum_{t=0}^{T-1}{\log(1 - D_{w_i}(s_t,a_t))}} + \frac{1}{|\mathcal{T}_E|} \frac{1}{T} \sum_{\tau \in \mathcal{T}_E} \sum_{t=0}^{T-1}{\log(D_{w_i}(s_t,a_t))} \right]
        \end{align*}
        \STATE Update policy parameters $\theta_i$ to $\theta_{i+1}$ as
        \begin{equation*}
            \quad \quad \quad \theta_{i+1} =  \argmax_{\theta} \left[ \frac{1}{|\mathcal{T}_G|} \frac{1}{T} \sum_{\tau \in \mathcal{T}_G} \sum_{t=0}^{T-1} \min \left( \frac{\pi_{\theta}(a_t | s_t)}{\pi_{\theta_i}(a_t | s_t)}A^i(s_t, a_t, r_t, s_{t+1}), g \left( \epsilon, A^i(s_t, a_t, r_t, s_{t+1}) \right) \right) \right]
        \end{equation*}
        \STATE Update value parameters $\phi_i$ to $\phi_{i+1}$ as 
        \begin{equation*}
            \quad \quad \quad \phi_{i+1} = \phi_{i} - \alpha \nabla_{\phi_i} \left[ \frac{1}{|\mathcal{T}_G|} \frac{1}{T} \sum_{\tau \in \mathcal{T}_G} \sum_{t=0}^{T-1} (r_t + V_{\phi_i}(s_{t+1}) - V_{\phi_i}(s_t))^2 \right]
        \end{equation*}
    \ENDIF
\ENDFOR
\end{algorithmic}
\end{algorithm*}

\section{Synthetic Data Generation}
\label{appendix:synth-data-gen}
Our synthetic data generation algorithm consists of three steps: \\
\textbf{(1) Pattern Injection and Correlation:} In this step, we take in a schema $\mathcal{S} = \{C_1, C_2 \dots C_k, N_1, N_2 \dots N_l, T_1, T_2 \dots T_m\}$ where $C_i, N_i, T_i$ represent Categorical, Numeric and Text columns respectively. For each column $c \in \mathcal{S}$, we randomly generate a set of \textit{patterns} $P(c) = \{p_1, p_2 \dots p_n\}$ and a random weighting $W(c) = \{w(p_1), w(p_2) \dots w(p_n)\}$ over these patterns ($\sum{w(p_i)} = 1$). These patterns are representative artifacts of the data appearing in column $c$, and the weights represent the likelihood that an element appearing in a column follows a pattern:
    \begin{itemize}
        \item For categorical columns, each $p_i$ is a category that appears in the column.
        \item For a numeric column, each $p_i$ is a tuple $(\mu_i, \sigma_i)$ representing a Gaussian distribution with mean and variance $(\mu_i, \sigma_i)$ respectively. Every element in the numeric column $N_j$ will be sampled from one of the $p_i \in P(N_j)$.
        \item For a text column, each $p_i$ is a tuple $(s_i, POS_i)$ where $s_i$ is a string and $POS_i \in \{START, MIDDLE, END\}$. If an element in a string column follows pattern $(s_i, POS_i)$, that element contains the string $s_i$ at the position specified by $POS_i$. The rest of it is padded with random strings.
    \end{itemize}

\begin{table}[htbp]
    \centering
    \footnotesize
    \begin{tabular}{ccc}
    \toprule
     \textbf{Dataset} & \textbf{Train} & \textbf{Evaluation} \\
     \midrule
    \textbf{1} & 614 & 154 \\
    \textbf{2} & 3686 & 922 \\
    \textbf{3} & 614 & 154 \\
    \textbf{4} & 921 & 231 \\
    \textbf{5} & 1638 & 410 \\
    \textbf{6} & 819 & 205 \\
    \textbf{7} & 1228 & 308 \\
    \bottomrule
    \end{tabular}
    \caption{Trajectory distribution for synthetic datasets}
    \label{tab:syn_dataset_distri}
\end{table}

We then randomly generate a list of dependencies of \textit{correlations} $\mathcal{C} = \{ \mathcal{C}_1, \mathcal{C}_2 \dots \mathcal{C}_n\}$ where each $\mathcal{C}_i = (c_{i_1}, c_{i_2}, \{r_{i_1}, r_{i_2} \dots r_{i_n}\})$ where every $c_{i_1}, c_{i_2} \in \mathcal{S}$ and $r_{i_k} = (p^{1}_{i_k}, p^{2}_{i_k})$ such that $p^{1}_{i_k} \in P(c_{i_1})$ and $p^{2}_{i_k} \in P(c_{i_2})$. Note that the term correlation here is not confused with mathematical correlations. Here, the term correlation between two columns indicates that the likelihood of a pattern appearing in one is influenced by the likelihood of patterns appearing in the other. Each $\mathcal{C}_i = (c_{i_1}, c_{i_2}, \{r_{i_1}, r_{i_2} \dots r_{i_n}\})$ is to be understood to mean that column $c_{i_1}$ is correlated with column $c_{i_2}$ in the following way: each $r_{i_k} = (p^{1}_{i_k}, p^{2}_{i_k})$ indicates that likelihood of $p^{2}_{i_k}$ appearing in column $c_{i_2}$ is greater every time $p^{1}_{i_k}$ appears in column $c_{i_1}$. We ensure that there are no cyclic chains formed in the list of correlations (a set $\{\mathcal{C}_1, \mathcal{C}_2 \dots \mathcal{C}_k\}$ such that $c_{1_2} = c_{2_1}, c_{2_2} = c_{3_1} \dots c_{n_2} = c_{1_1}$ ) and limit the number of correlations that a column can appear in to make sure the set of correlations forms a Directed Acyclic Graph (DAG) over the set of columns. 

\noindent
\textbf{(2) Row population:} In this step, we take the schema $\mathcal{S}$ and generated list of correlations $\mathcal{C}$ and generate the rows of the dataset. We generate each row one by one. To generate the element at column $c$ in a row, we do the following:
    \begin{itemize}
        \item Set base distribution $W = W(c)$ and get the relevant set of correlations $C_R = \{\mathcal{C}_i \in \mathcal{C} \text{ such that } \ c_{i_2} = c\}$.
        \item For each correlation $\mathcal{C}_i \in \mathcal{C}_R$, for each $r_{i_k} = (p^{1}_{i_k}, p^{2}_{i_k})$, multiply weight of $p^{2}_{i_k}$ in $W$ with pre-defined multiplier $m$.
        \item Normalize $W$ and sample pattern $p$ according to updated weights. Generate elements according to $p$ as described above.
    \end{itemize}

\textbf{(3) Trajectory Generation:} The correlations generated in step 1 form a Directed Acyclic Graph (DAG) where the nodes are columns and the edges correspond to the correlation between them. We model an example of an EDA trajectory to be a traversal of this DAG from the root node (corresponding to the base view) to each leaf node and back. The path taken determines the correlations uncovered in the analysis session with each edge traversal corresponding to a particular analysis action. The column type and its patterns determine the operations performed during this process. For instance, if there is an edge between two categorical columns ($c_i$ and $c_j$) with related patterns values $p_i$ and $p_j$, the possible order of operation would be either [\texttt{FILTER $c_i$ EQ $p_i$}, \texttt{FILTER $c_j$ EQ $p_j$}] or [\texttt{FILTER $p_i$ EQ $v_j$}, \texttt{GROUP $c_i$ AGGREGATE COUNT $c_j$}]. Numerical columns are filtered based on the mean of the distribution from which the values in each column are sampled, and text columns are filtered based on the pattern present in the column and its position. The algorithm returns to the root node of the subtree using \texttt{BACK} actions, and the trajectory ends when each node has been visited in topological order. We traverse the graph in a depth-first search approach to generate expert trajectories over the synthetically generated dataset. Each trajectory forms an expert EDA session.

\end{document}